\documentclass[letterpaper, 10 pt, conference]{ieeeconf}

\IEEEoverridecommandlockouts
\usepackage{graphicx}
\usepackage[table,xcdraw]{xcolor}
\usepackage{amsmath}
\usepackage{amssymb}
\usepackage{bm}
\usepackage{multirow}
\usepackage{cuted}
\usepackage{stmaryrd}
\usepackage{booktabs}
\definecolor{cvprblue}{rgb}{0.21,0.49,0.74}

\usepackage[pagebackref,breaklinks,colorlinks,allcolors=cvprblue]{hyperref}
\usepackage{hyperref}
\usepackage{cleveref}

\usepackage{stmaryrd}

\definecolor{secondcolor}{RGB}{180,198,231}
\definecolor{firstcolor}{RGB}{240,220,220}

\definecolor{conical}{HTML}{D64545}     
\definecolor{round-flask}{HTML}{2E8B57} 
\definecolor{wine}{HTML}{AED4E5}        
\definecolor{measuring}{HTML}{6EA3C1}   
\definecolor{new}{HTML}{C3C769}         
\definecolor{bedC}{HTML}{FFCC99}        
\definecolor{table}{HTML}{7F5AA2}       
\definecolor{tableC}{HTML}{2F69A3}      
\definecolor{beaker}{HTML}{9BCB3B}      
\definecolor{cup}{HTML}{FF8C3A}         
\definecolor{con}{HTML}{B7A7D8}         

\newcommand{\legendcell}[2]{%
  \begin{tabular}{@{}c@{}}
    {\color{#1}\rule{10pt}{10pt}}\\[-2pt]%
    \rotatebox{90}{#2}%
  \end{tabular}%
}
\title{\LARGE \bf
Trans2Occ: Voxel Occupancy Estimation and Grasp for Transparent Objects from Simulation to Reality
}

\author{
    Yixuan Yang$^{1,2,*}$\quad
    Sha Zhang$^{3,*}$\quad
    Rui Li$^{1,4,*}$\quad
    Zhenfei Yin$^{5}$\quad
    Xinzhu Ma$^{1,6}$\\
    Yiran Qin$^{5}$\quad
    Lei Bai$^{1,3}$\quad
    Xudong Xu$^{1}$\quad
    Shilin Shan$^{7}$\quad
    Wangmeng Zuo$^{4}$\\
    Yanyong Zhang$^{8}$\quad
    Wanli Ouyang$^{1,3}$\quad
    Feng Zheng$^{2,\dagger}$\quad
    Shixiang Tang$^{1,\dagger}$\quad
    Dongzhan Zhou$^{1,\dagger}$\\[1mm]
    $^{1}$Shanghai AI Laboratory \quad
    $^{2}$SUSTech \quad
    $^{3}$CUHK \quad
    $^{4}$Harbin Institute of Technology\\
    $^{5}$University of Oxford \quad
    $^{6}$Beihang University\\
    $^{7}$Nanyang Technological University \quad
    $^{8}$University of Science and Technology of China\\[1mm]
    \tt \small
    arnoldyang97@gmail.com,\quad zhangsha2048@gmail.com,\quad lirui.work0@gmail.com
}

\begin{document}

\maketitle

\maketitle

\begin{abstract}

Transparent objects remain challenging for robotic perception due to unreliable depth sensing caused by refraction and reflection. While prior approaches rely on multi-view reconstruction or depth completion, they are often difficult to scale or deploy in real-world robotic systems.
In this paper, we present a practical framework for transparent object perception and manipulation based on single-view RGB input. Our approach predicts voxel-space occupancy directly from a single image, providing a geometry-aware representation that supports downstream robotic grasping. To enable large-scale training, we construct a simulation pipeline that generates paired RGB images and voxel occupancy annotations under diverse materials and lighting conditions.
We demonstrate that the predicted occupancy representation is robust to domain shifts and transfers effectively from simulation to real-world robotic setups without fine-tuning. A simple rule-based grasping strategy built on top of the occupancy further achieves reliable grasp performance on transparent objects.
Extensive experiments in both simulation and real-world environments show that our framework provides accurate 3D understanding and enables practical manipulation of transparent objects. These results suggest that single-view occupancy prediction offers a scalable and effective solution for transparent object perception in robotics.

\end{abstract}    

\section{Introduction}
\label{sec:intro}

Transparent objects are widely encountered in everyday environments and scientific laboratories, including glassware, plastic containers, and chemical apparatus. Enabling robots to reliably manipulate such objects is important for applications such as laboratory automation and industrial handling~\cite{li2025labutopia, seifrid2022autonomous, lan2025autobio, li2024chemistry3d}. However, transparent objects remain particularly challenging for robotic perception systems, as standard depth sensors often fail due to refraction and reflection, resulting in incomplete or inaccurate geometric observations~\cite{xu2021seeing, sajjan2020clear, chen2022clearpose}.

\begin{figure}[t]
    \centering
    \includegraphics[width=1.0\linewidth]{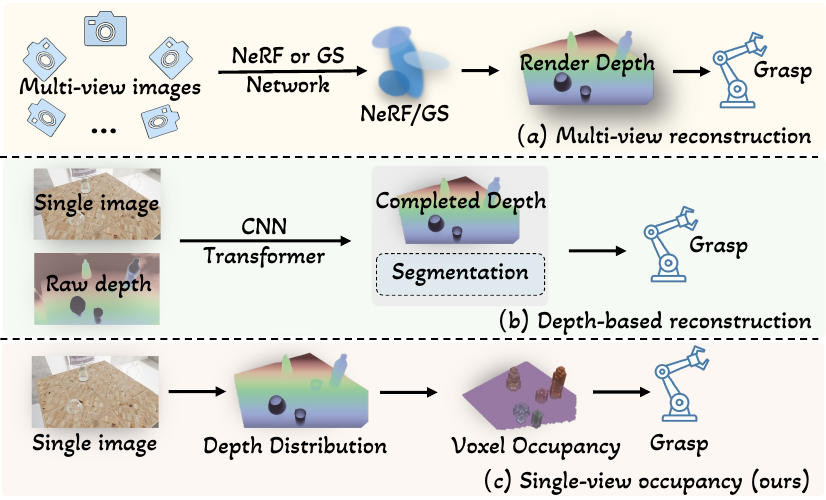}
\caption{Comparison of representative paradigms for transparent object perception.
(a) Multi-view reconstruction methods rely on multiple observations and are difficult to deploy in real-time robotic settings.
(b) Depth-based approaches depend on sensor measurements, which are often unreliable for transparent materials.
(c) Our approach predicts voxel occupancy from a single RGB image, providing a compact and geometry-consistent representation for downstream manipulation.
The figure highlights the trade-off between reconstruction fidelity and practical deployability in robotic systems.}
    \label{fig:teaser}
    \vspace{-2em}
\end{figure}

Existing approaches for transparent object reconstruction typically rely on either multi-view observations or depth-based estimation. Multi-view methods, such as neural rendering techniques~\cite{ichnowski2021dex, duisterhof2024residual, kerr2022evo, fujitomi2022lb, chen2023nerrf, li2025tsgs, huang2025transparentgs, kim2025transplat, agrawal2024clear}, require multiple calibrated viewpoints, which are difficult to obtain in real-time robotic settings. Depth-based approaches~\cite{yang2024depth, fang2022transcg, liu2025monocular, costanzino2023learning, chen2022tode, meng2024dfnet} attempt to complete missing depth measurements, but transparent materials often lead to unreliable or corrupted sensor readings, making accurate supervision and generalization challenging.
As illustrated in Fig.~\ref{fig:teaser}, these approaches face a fundamental trade-off between reconstruction fidelity and deployability. Multi-view methods are difficult to integrate into real-time robotic systems, while depth-based methods suffer from unreliable sensing. These limitations hinder their practical use in robotic manipulation tasks.

To address these limitations, we explore an alternative formulation by directly predicting voxel occupancy from a single RGB image. The key idea is to infer a coarse but geometry-consistent 3D representation that is sufficient for downstream manipulation tasks. Compared to dense surface reconstruction, voxel occupancy provides a compact and structured representation of object location and shape, which naturally aligns with robotic grasp planning.

This formulation offers two key advantages. First, voxel occupancy captures coarse geometry that is sufficient for manipulation, without requiring precise surface reconstruction. Second, it is naturally compatible with simulation-based training. RGB images exhibit a smaller domain gap between simulation and the real world compared to depth measurements, making it feasible to generate large-scale training data in simulation. By learning a mapping from RGB observations to voxel occupancy, the resulting representation becomes more robust to domain variations and can be transferred to real robotic setups without requiring additional adaptation.

Based on this insight, we present Trans2Occ, a practical framework that predicts voxel occupancy from a single RGB image to support downstream robotic grasping. To enable scalable training, we construct a simulation pipeline that generates paired RGB images and voxel occupancy annotations under diverse conditions. A geometry-aware model infers occupancy from the input image, and a simple rule-based strategy operates on the resulting representation, demonstrating that occupancy alone provides sufficient geometric information for reliable manipulation.

We evaluate our approach in both simulation and real-world robotic environments. Experimental results demonstrate that the proposed framework achieves accurate occupancy prediction and enables reliable grasping of transparent objects. Notably, models trained purely on simulated data generalize effectively to real-world setups without fine-tuning, highlighting the practicality of the proposed representation for robotic applications.

Our contributions are summarized as follows:

\begin{itemize}
    \item We propose a practical single-view framework for transparent object grasping that leverages voxel occupancy as a geometry-aware representation inferred directly from RGB images, without relying on depth sensing or multi-view observations.
    \item We develop a simulation-driven data generation pipeline that provides scalable supervision for transparent object perception, supporting effective sim-to-real transfer.
    \item We provide empirical and qualitative analysis demonstrating that voxel occupancy serves as a robust and geometry-consistent representation for transparent objects in robotic manipulation tasks.
\end{itemize}

\section{Related Work}

\begin{figure*}[th]
    \centering
    \includegraphics[width=0.9\linewidth]{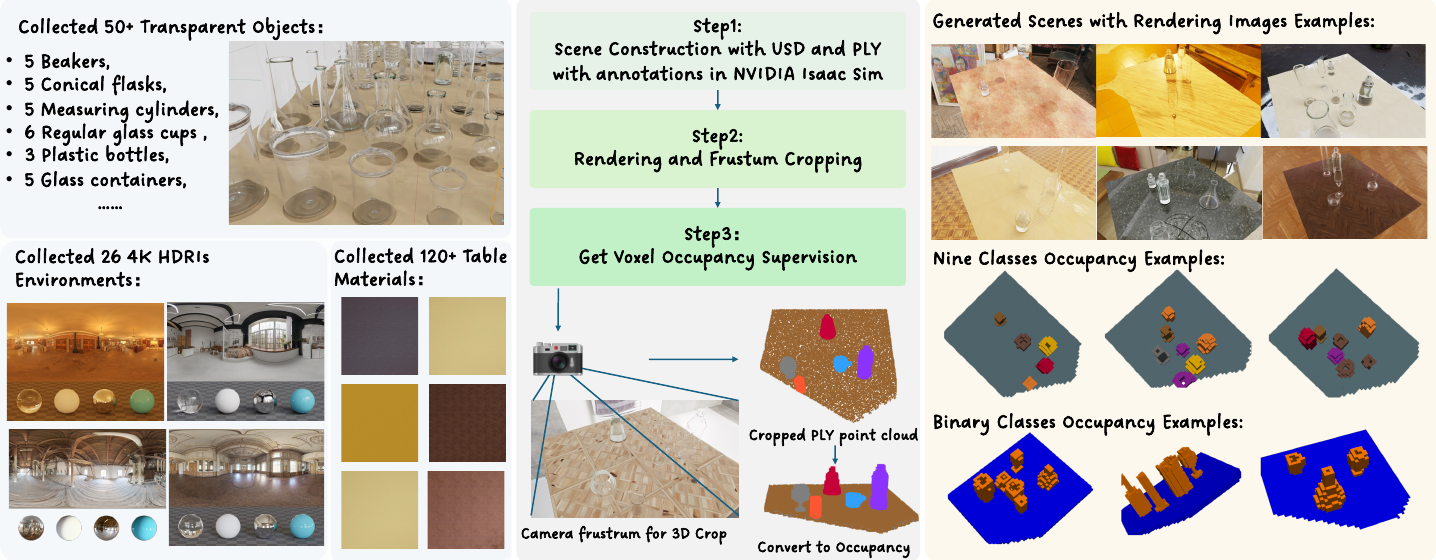}
    \caption{\textbf{Overview of the Sim-Trans3D data generation pipeline.}
The pipeline generates paired RGB observations and voxel occupancy labels for transparent objects in a fully controlled environment. Diverse tabletop scenes are constructed under varying materials, lighting conditions, and object configurations. Using known camera parameters, each scene is rendered from a fixed viewpoint, and the visible region is extracted via the camera frustum. The corresponding 3D geometry is discretized into voxel-space occupancy aligned with the input image, establishing a direct mapping from 2D observations to 3D spatial structure. This design avoids intermediate representations such as depth and enables scalable, simulation-driven supervision for transparent object perception.}
    \label{fig:data-architecture}
\end{figure*}

\subsection{Perception of Transparent Objects}
Perception of transparent objects remains a significant challenge in robotic vision, primarily due to low-contrast textures and complex optical phenomena such as refraction and reflection, which make it difficult for conventional sensors to accurately recover geometry and depth~\cite{jiang2023robotic,xie2020segmenting}. While RGB-D cameras can provide stable geometric information for opaque objects, they often fail when applied to transparent ones. To address this issue, various approaches have been proposed. Polarization cameras can suppress specular highlights and enhance edge features~\cite{shao2024polarimetric,kalra2020deep,mei2022glass}, but their high cost limits practical deployment. Sim2Real and synthetic dataset-based methods leverage controllable environments to train models, improving segmentation and tracking performance for transparent objects~\cite{wang2025transdiff,bai2024close,mei2022glass,sajjan2020clear,yu2023tgf}. Moreover, multi-task perception networks attempt to jointly learn segmentation, depth, and pose to support manipulation tasks~\cite{wang2023mvtrans}. Neural rendering methods, such as NeRF, and volumetric reconstruction have shown potential in capturing complex light propagation~\cite{kerr2022evo,ichnowski2021dex,li2020through}, while stereo and multi-view techniques exploit geometric consistency to enhance depth estimation accuracy~\cite{fang2022transcg,bai2024cleardepth,li2023fdct}. However, most of these methods rely on multi-sensor setups or strictly calibrated camera systems, limiting their adoption on lightweight robotic platforms. In contrast, we propose a monocular RGB-based method for transparent object reconstruction, enabling shape recovery from a single image.

\subsection{Transparent Object Datasets}
Real-world transparent object datasets typically contain multi-modal sensory data, including RGB images~\cite{fang2022transcg,bai2024cleardepth,sajjan2020clear,zhu2021rgb,xu2021seeing}, polarization images~\cite{mei2022glass}, and thermal images~\cite{huo2023glass}. These datasets often rely on labor-intensive pixel-level annotation of glass regions. However, recent studies have shown that real-world datasets tend to suffer from label noise and inconsistencies in acquisition conditions, which can lead to degraded model performance~\cite{yang2024depth,bai2024cleardepth}. In contrast, high-quality synthetic datasets with precise annotations have been demonstrated to significantly improve model generalization and stability. Nevertheless, high-fidelity synthetic datasets that span multiple data domains remain scarce. Existing works often use physically based ray-tracing renderers to narrow the sim-to-real gap, but domain discrepancies still persist and continue to limit performance.
Therefore, constructing high-quality, controllable, and diverse synthetic datasets for transparent objects remains of great importance~\cite{xu2021seeing,bai2024cleardepth,dai2022domain,shi2024asgrasp,chen2018tom}. Furthermore, many of these datasets rely on cumbersome pre-/post-processing steps (such as image segmentation and background reconstruction), offering limited support for end-to-end embodied intelligence algorithms. Although some simulation datasets tailored for specific sensors attempt to reproduce the generation process of RGB~\cite{dai2022domain,shi2024asgrasp}, infrared projection, and noisy depth data, their rendering pipelines are often complex and computationally inefficient, making large-scale expansion difficult. 
In contrast, our dataset provides a clean and well-structured design with precise annotations. 
It includes complete 3D object and scene models, together with 3D semantic masks and voxel occupancy for every rendered viewpoint.

Despite these advances, existing methods either rely on multi-view observations or depend on unreliable depth signals, making them difficult to deploy in practical robotic systems. In contrast, our approach focuses on learning a direct mapping from RGB images to voxel occupancy, enabling robust and scalable perception under real-world constraints.

\section{Method}

Our goal is to recover the 3D structure of transparent objects from a single RGB image for robotic manipulation. Due to the ambiguity introduced by refraction and reflection, precise surface reconstruction is often unreliable for transparent materials. Instead, we focus on estimating a coarse yet structured representation that captures the spatial extent of objects and is sufficient for downstream grasp planning.

\begin{figure*}[t]
    \centering
    \includegraphics[width=0.9\linewidth]{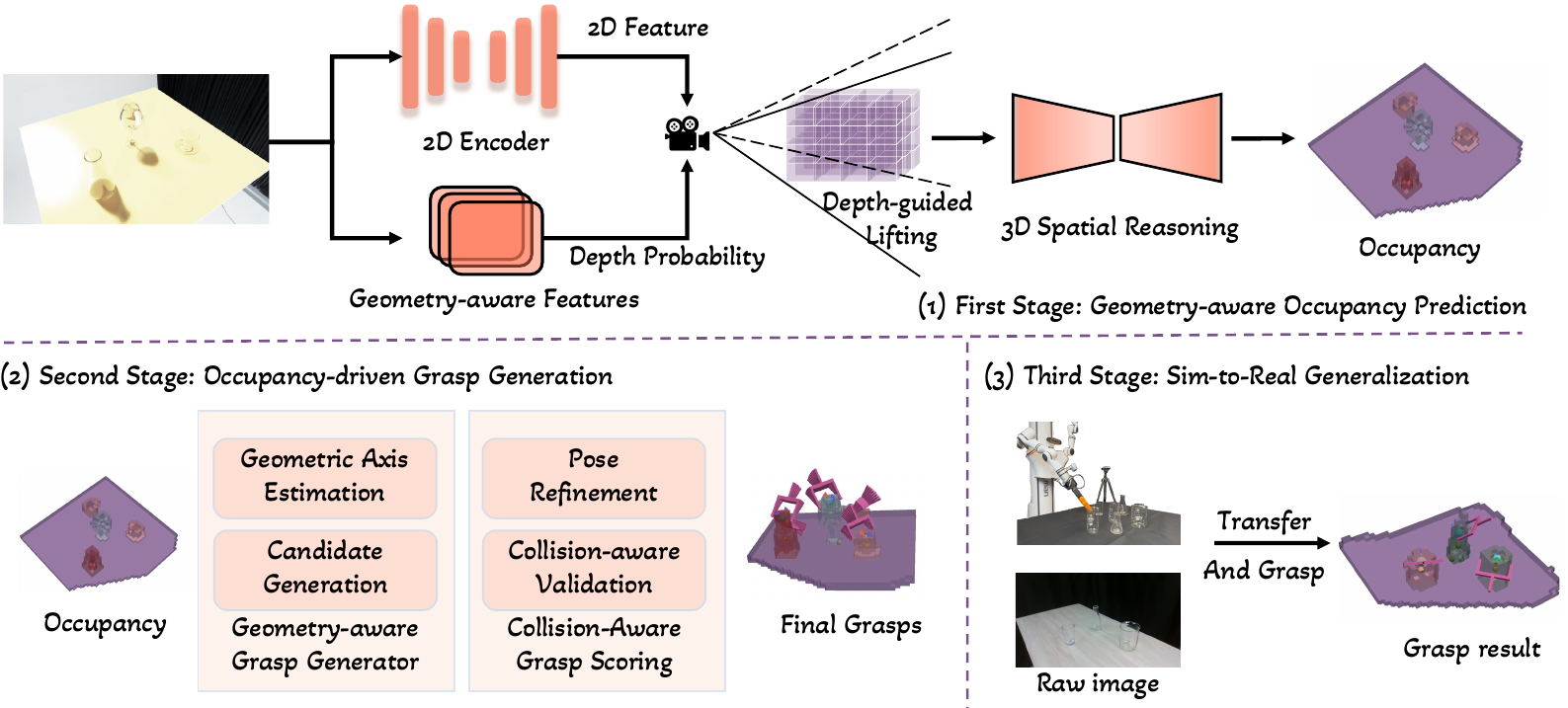}
    \caption{\textbf{Overview of the Trans2Occ framework.}
    The framework consists of three stages.
    (1) \textit{Geometry-aware occupancy prediction:} Given a single RGB image, semantic features and geometry-aware cues are extracted and lifted into a voxel-aligned 3D representation via depth-guided projection, followed by 3D spatial reasoning to predict occupancy.
    (2) \textit{Occupancy-driven grasp generation:} The predicted occupancy is used to infer feasible grasp poses by analyzing spatial structure and ensuring collision-aware placement.
    (3) \textit{Sim-to-real transfer:} The model trained in simulation generalizes to real-world robotic setups without additional adaptation, enabling reliable grasping of transparent objects.}
    \label{fig:trans2occ-architecture}
    \vspace{-1em}
\end{figure*}

We formulate this task as voxel occupancy prediction, where the model infers a discretized 3D grid encoding object presence in space. This representation balances geometric expressiveness and robustness, and provides a convenient interface for reasoning and action in robotic systems.
Based on this formulation, our framework consists of three components: (1) a simulation pipeline for generating large-scale training data, (2) a geometry-aware model for predicting voxel occupancy from a single image, and (3) a lightweight grasping strategy that operates on the predicted occupancy. An overview of the framework is shown in Fig.~\ref{fig:trans2occ-architecture}.

\subsection{Simulation Pipeline}
To enable scalable supervision for transparent object perception, we develop a simulation-based data generation pipeline, termed \textbf{Sim-Trans3D}. As illustrated in Fig.~\ref{fig:data-architecture}, the pipeline generates paired RGB observations and voxel occupancy labels in a fully controlled environment, addressing the difficulty of obtaining reliable 3D annotations for transparent objects in real-world settings.

The pipeline follows a structured process that links 2D observations with 3D spatial representations. First, diverse tabletop scenes are constructed with transparent objects under varying materials, lighting conditions, and object configurations, ensuring variability in both appearance and geometry. Second, given known camera intrinsics and extrinsics, each scene is rendered from a fixed viewpoint, and the visible region is extracted using the camera frustum. This step establishes a consistent correspondence between image observations and the underlying 3D structure. Third, the reconstructed 3D geometry is discretized into a voxel grid aligned with the camera frame, producing occupancy representations that serve as direct supervision.

Compared to conventional datasets that rely on depth or segmentation annotations, Sim-Trans3D directly provides voxel-level supervision that is consistent with the target representation of our model. This design avoids intermediate representations such as depth completion or multi-view reconstruction, and enables learning a direct mapping from RGB images to 3D occupancy. The resulting dataset supports large-scale training and improves robustness to domain variations, facilitating effective transfer from simulation to real-world robotic environments.

For clarity, we summarize the main stages of the pipeline below:

\textbf{(1) Asset Preparation.} We collect diverse transparent objects (e.g., beakers, flasks, and containers) together with environment assets, and align them to a consistent coordinate frame.
\textbf{(2) Scene Generation.} We synthesize tabletop scenes with randomized object layouts, materials, and illumination conditions, and record object poses for each configuration.
\textbf{(3) Camera-Posed Rendering and Frustum Cropping.} Scenes are rendered from a fixed camera setup, and the camera frustum is used to extract visible regions, producing RGB images and corresponding 3D observations.
\textbf{(4) Voxelization.} The cropped 3D geometry is discretized into a voxel grid aligned with the camera frame, yielding occupancy labels (and optional semantic annotations) for training.

\subsection{Occupancy Prediction Model}

Given a single RGB image, our goal is to infer a voxel occupancy representation that captures the 3D spatial structure of transparent objects for robotic manipulation. This task is particularly challenging due to the ambiguity introduced by refraction and reflection, which makes reliable depth estimation difficult.
Instead of relying on raw depth measurements, we adopt a geometry-aware lifting formulation that directly maps 2D image observations into a structured 3D representation. The key idea is to infer a probabilistic correspondence between image pixels and spatial locations, allowing the model to reason about geometry under uncertainty.
Based on this formulation, the model progressively transforms image features into a voxel-aligned representation through three steps: (1) extracting dense 2D features, (2) lifting them into 3D space via a depth-guided projection, and (3) performing spatial reasoning in volumetric space to predict occupancy.
The overall framework is illustrated in Fig.~\ref{fig:trans2occ-architecture}.

\paragraph{Formalization.}
Given an input RGB image $\mathbf{I} \in \mathbb{R}^{H \times W \times 3}$, our goal is to predict a 3D voxel occupancy grid representing the spatial structure of the scene. We extract dense 2D features $\mathbf{F}_{2D} \in \mathbb{R}^{H \times W \times C}$ using a pre-trained image encoder $\mathcal{N}_{2D}(\cdot)$, and geometry-aware features $\mathbf{F}_{3D} \in \mathbb{R}^{H \times W \times M}$ using a vision transformer backbone.

To establish correspondence between 2D observations and 3D space, we discretize the depth range into $K$ bins and predict a per-pixel depth probability distribution $P(u, v, d_i)$ over each bin $d_i$. Given camera intrinsics $\mathbf{K}$, the 3D point corresponding to depth $d_i$ at pixel $(u, v)$ is computed as:
$$
\mathbf{X}(u, v, d_i) = d_i \cdot \mathbf{K}^{-1} [u, v, 1]^T.
$$

We then lift 2D features into a 3D voxel space by aggregating contributions from all pixels and depth bins. The feature at voxel $(x, y, z)$ is defined as:
$$
\mathbf{F}_{xyz} = \sum_{(u,v), i} P(u, v, d_i) \cdot \mathbf{f}_{2D}(u, v),
$$
where the summation is taken over all pixels whose projected rays pass through the voxel.

This formulation converts depth-aware 2D features into a voxel-aligned 3D representation, enabling spatial reasoning in volumetric space.

\paragraph{3D Spatial Reasoning and Prediction.}
The lifted features are processed using a 3D convolutional network to hierarchically encode and aggregate spatial context. An occupancy head then predicts voxel-wise foreground/background probabilities as well as semantic labels, producing the final voxel-space representation.

\paragraph{Training Objective.}
The training objective consists of three components: the occupancy loss $\mathcal{L}_{occ}$, the semantic classification loss $\mathcal{L}_{sem}$, and the depth estimation loss $\mathcal{L}_{dep}$. The overall loss is formulated as:
$$
\mathcal{L} = \lambda_{occ} \mathcal{L}_{occ} + \lambda_{sem} \mathcal{L}_{sem} + \lambda_{dep} \mathcal{L}_{dep},
$$
where $\lambda_{occ}$, $\lambda_{sem}$, and $\lambda_{dep}$ are balancing coefficients. $\mathcal{L}_{occ}$ supervises occupancy prediction, $\mathcal{L}_{sem}$ is a cross-entropy loss for semantic classification, and $\mathcal{L}_{dep}$ is a binary cross-entropy loss applied to depth prediction.

The predicted occupancy representation serves as the geometric foundation for downstream manipulation tasks, as described in the next section.

\subsection{Occupancy-Driven Grasp Pose Generation}

Given the predicted voxel occupancy, our goal is to generate feasible grasp poses for transparent objects. Rather than relying on a learned grasping network, we adopt a geometry-driven strategy that operates directly on the occupancy representation. This design allows us to explicitly evaluate whether the predicted 3D structure is sufficient for downstream manipulation.

We first extract object regions from the occupancy grid and estimate their principal geometric axes using shape analysis. Based on the inferred spatial structure, candidate grasp poses are generated by sampling approach directions aligned with the object geometry. Each candidate is further refined by adjusting its position and orientation to better match local surface structure.
To ensure physical feasibility, we perform collision-aware validation in the voxel space. Candidate grasps are evaluated based on geometric consistency and clearance, and the highest-scoring collision-free pose is selected as the final grasp.
This pipeline demonstrates that the predicted occupancy representation provides sufficient geometric information for reliable grasp generation, without requiring additional learning-based grasp models.

\subsection{Sim-to-Real Transfer}

To evaluate the transferability of the proposed framework, we deploy the trained model directly in a real-world robotic setup. A single RGB image is captured as input, and the model predicts voxel occupancy and semantic information, which are subsequently used for grasp generation.
Importantly, no additional adaptation techniques such as domain adaptation, style transfer, or real-world fine-tuning are applied. Despite the domain gap between simulation and real environments, the system achieves reliable grasp performance.
This result highlights that voxel occupancy serves as a robust and domain-invariant representation, enabling effective transfer of transparent object perception and manipulation from simulation to real-world scenarios.

\section{Experiments}

\begin{table*}[th]
\centering
\caption{Per-class IoU (\%) and mIoU on the benchmark. Best numbers are in \colorbox{secondcolor}{blue}.}
\resizebox{\linewidth}{!}{%
\begin{tabular}{l | c c| *{9}{c} c | c c c}
\toprule
Method & Input & IoU
& \legendcell{beaker}{Beaker}
& \legendcell{conical}{Conical}
& \legendcell{measuring}{Cylinder}
& \legendcell{cup}{Cup}
& \legendcell{wine}{Wine}
& \legendcell{bedC}{Bottle}
& \legendcell{con}{Container}
& \legendcell{round-flask}{Flask}
& \legendcell{table}{Tabletop}
& mIoU
& \legendcell{measuring}{Transparent} & \legendcell{conical}{Tabletop}&mIoU \\
\midrule
MonoScene~\cite{cao2022monoscene} & $x^{\text{rgb}}$
& 79.44
& 36.25 & 26.21 & 36.94 & 42.47 & 33.20 & 47.62 & 24.05 & 26.21 &  94.39
& 43.36 & 40.37 & 94.39 &67.38\\
ISO~\cite{yu2024monocular} & $x^{\text{rgb}}$
& 81.45
& 44.13 & 29.42 &\cellcolor[rgb]{ .706,  .776,  .906} 40.90 & 35.61 & 36.81 & 47.38 & 24.48 & 28.32 & 95.01
& 45.83 & 44.25 & 95.01 & 69.63 \\
\midrule
Trans2Occ (Ours) & $x^{\text{rgb}}$
& \cellcolor[rgb]{ .706,  .776,  .906}81.60
& \cellcolor[rgb]{ .706,  .776,  .906}44.74 & \cellcolor[rgb]{ .706,  .776,  .906}31.31 & 39.57 &\cellcolor[rgb]{ .706,  .776,  .906} 42.57 & \cellcolor[rgb]{ .706,  .776,  .906}38.41 &\cellcolor[rgb]{ .706,  .776,  .906} 49.89 &\cellcolor[rgb]{ .706,  .776,  .906} 30.95 &\cellcolor[rgb]{ .706,  .776,  .906} 31.31 &\cellcolor[rgb]{ .706,  .776,  .906} 95.39
& \cellcolor[rgb]{ .706,  .776,  .906}46.90 &\cellcolor[rgb]{ .706,  .776,  .906} 46.92 &\cellcolor[rgb]{ .706,  .776,  .906} 98.80 &\cellcolor[rgb]{ .706,  .776,  .906}72.32 \\
\bottomrule
\end{tabular}}
\label{tab:perclass_iou}
\end{table*}

\begin{figure}[t]
    \centering
    \includegraphics[width=1\linewidth]{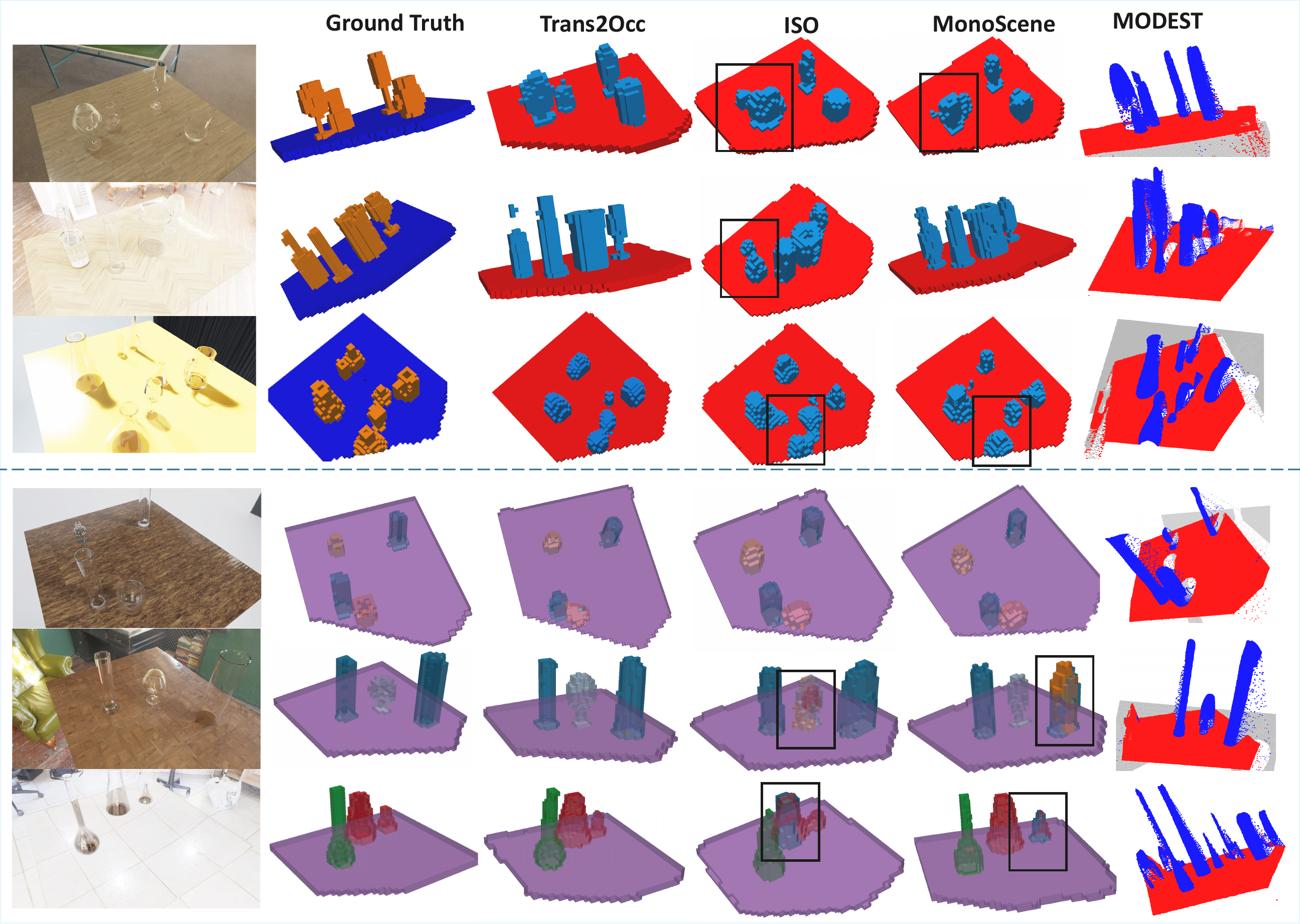}
    \vspace{-1em}
    \caption{Qualitative comparison of multi-class and binary-class occupancy prediction. Errors in black boxes. Please zoom in to see the detail.
    }
    \vspace{-1em}
    \label{fig:occ_results}
\end{figure}
\subsection{Experiments Settings}
\noindent \textbf{Dataset Construction}
We build the training data entirely with our Sim-Trans3D pipeline. We first randomize illumination environments and desktop materials to synthesize 100 scene backgrounds. For each background, we generate 50 placement configurations of transparent objects, resulting in 5,000 samples in total (100 × 50). During rendering, we also randomize lighting intensity per image. For each sample, we store the RGB image, the corresponding depth map, and the voxel-space occupancy ground truth (fully annotated) as the final supervision signal.
We apply 4,800 data samples for training and 200 for validation. In addition, we synthesize 10 unseen scene backgrounds and produce 100 samples for testing. Beyond simulation, we collect 40 real-world samples by mounting an Intel D435i camera in diverse lighting conditions, which are used exclusively for Sim-to-Real real-world testing.

\noindent \textbf{Implementation Details}
We restrict the workspace to a tabletop volume of \(1.0\,\mathrm{m}\times 1.0\,\mathrm{m}\times 0.5\,\mathrm{m}\), discretized into a \(64\times 64\times 32\) voxel grid. An Intel RealSense D435i camera is rigidly mounted on the left side of the table with fixed extrinsics both in simulation and the real environment. 
We train Trans2Occ in both multiclass and binary settings under the same configuration for \(70\) epochs with a total batch size of \(72\) (\(9\) per GPU on \(8\) GPUs), using AdamW with a base learning rate of \(1\times 10^{-4}\) and weight decay of \(5\times 10^{-4}\). The learning rate follows cosine annealing with warm restarts \(\bigl(T_{0}=10,\ T_{\mathrm{mult}}=2,\ \eta_{\min}=10^{-7}\bigr)\). Parameters of the depth sub-network use \(0.1\times\) the base learning rate. The loss is a weighted sum of multiple components with uncertainty-based automatic balancing; base weights are \(1.0\) for all terms (cross-entropy, semantic, and depth-geometric losses). All experiments are conducted on \(8\times\) NVIDIA A100 GPUs.

We consider two evaluation settings. (i) Binary occupancy: the model distinguishes only between transparent objects and the tabletop, without categorizing object types. (ii) Diverse (multiclass) occupancy: the model additionally assigns a category label to each transparent object among {beaker, conical flask, round-bottomed flask, measuring cylinder, regular glass cup, wine glass cup, common plastic bottle, glass container, tabletop}. In the experiments, we denote these settings as Trans2Occ-BC (binary classification) and Trans2Occ-DC (diverse classification).

\noindent \textbf{Evaluation Metrics}
\label{sec:metrics}
We evaluate our method from three perspectives: depth estimation, occupancy prediction, and grasping performance. For depth and occupancy, we adopt standard metrics such as MAE, RMSE, REL, IoU, and mIoU to measure geometric fidelity in both binary and semantic settings. For grasping, we report success rates in simulation and in real-world trials, where a grasp is counted as successful if the object is lifted stably without slippage or collision. Detailed metric definitions and implementation details are provided in the supplementary material.

\subsection{Compared Methods}
We compare Trans2Occ with representative methods from three related areas, all evaluated under our single-view setting. For depth estimation, we use monocular baselines DepthAnythingV2~\cite{yang2024depthv2} and Marigold~\cite{ke2023repurposing}; NeRF-style methods~\cite{duisterhof2024residual, ichnowski2021dex} typically require many multi-view images to reconstruct a scene and are therefore not applicable here. For occupancy prediction, we compare with recent voxel-based and transformer-based methods MonoScene~\cite{cao2022monoscene} and ISO~\cite{yu2024monocular}, which we train or fine-tune on our simulation-generated dataset when possible. For transparent-object grasping, we include AnyGrasp~\cite{fang2023anygrasp} and MODEST~\cite{liu2025monocular}. In contrast to all these baselines, Trans2Occ jointly predicts volumetric occupancy and semantic categories directly from a single RGB image, providing a compact 3D representation that can be used for grasp planning.

\subsection{Experiments Results}

We evaluate the proposed framework in both simulation and real-world environments from multiple perspectives, including depth reconstruction, occupancy prediction, and grasping performance. 
All metrics follow the definitions introduced in Sec.~\ref{sec:metrics}. 
Unless otherwise stated, the model is trained on the Sim-Trans3D dataset.

\subsubsection{In Simulation Environment}
\noindent \textbf{Auxiliary Depth Estimation Analysis}
We analyze the auxiliary depth prediction to better understand the learned geometry.
The predicted depth maps are projected back into the camera frustum and compared with the simulated ground truth. 
As shown in ~\cref{tab:depth_metrics}, Trans2Occ produces geometry-consistent depth estimates and higher consistency across object boundaries compared with all baseline methods. 
Notably, it reconstructs fine geometric details around transparent regions where conventional depth networks often fail.

\begin{table}[t]
    \centering
    \caption{Depth Estimation Metrics Comparison (Normalized)}
    \label{tab:depth_metrics}
    \resizebox{1\linewidth}{!}{
    \begin{tabular}{lcccccc}
        \toprule
        \textbf{Methods} & \multicolumn{6}{c}{\textbf{Metrics}} \\
        \cmidrule(lr){2-7}
        & RMSE $\downarrow$ & REL $\downarrow$ & MAE $\downarrow$ & $\delta_{1.05}$ $\uparrow$ & $\delta_{1.10}$ $\uparrow$ & $\delta_{1.25}$ $\uparrow$ \\
        \midrule
        DepthAnything2 &  0.7383 &  0.7856 & 1.2317 & 0.00 & 0.00 & 0.00 \\
        Marigold &  0.4219 &  0.6898 & 1.7253 & 0.22 & 0.39 & 1.82 \\
        Trans2Occ (VGGT) &\cellcolor[rgb]{ .706,  .776,  .906} 0.3563 &\cellcolor[rgb]{ .706,  .776,  .906} 0.6126 &\cellcolor[rgb]{ .706,  .776,  .906} 0.3515 &\cellcolor[rgb]{ .706,  .776,  .906} 0.25 &\cellcolor[rgb]{ .706,  .776,  .906} 0.53 &\cellcolor[rgb]{ .706,  .776,  .906} 3.85 \\
        
        \bottomrule
    \end{tabular}}
    \vspace{-1.5em}
\end{table}

\noindent \textbf{Occupancy Prediction}
We further evaluate voxel-level occupancy prediction in both binary and semantic settings. 
As shown in Table~\ref{tab:perclass_iou}, compared with recent voxel-based methods, Trans2Occ yields higher IoU in almost all classes and  mIoU, especially near transparent boundaries.
The model demonstrates strong spatial awareness and accurate semantic separation between overlapping transparent objects.

\noindent \textbf{Grasp in Simulation}
To validate whether the predicted occupancy representation supports manipulation, we conduct grasp experiments using the simulated robotic arm. 
Candidate grasps are sampled from occupied voxels and refined through collision checking before execution. 
Our approach achieves over 71.67\% grasp success across multiple object categories, showing that voxelized occupancy provides a stable geometric foundation for grasp planning.

\begin{figure}[t] 
    \centering
    \includegraphics[width=1\columnwidth]{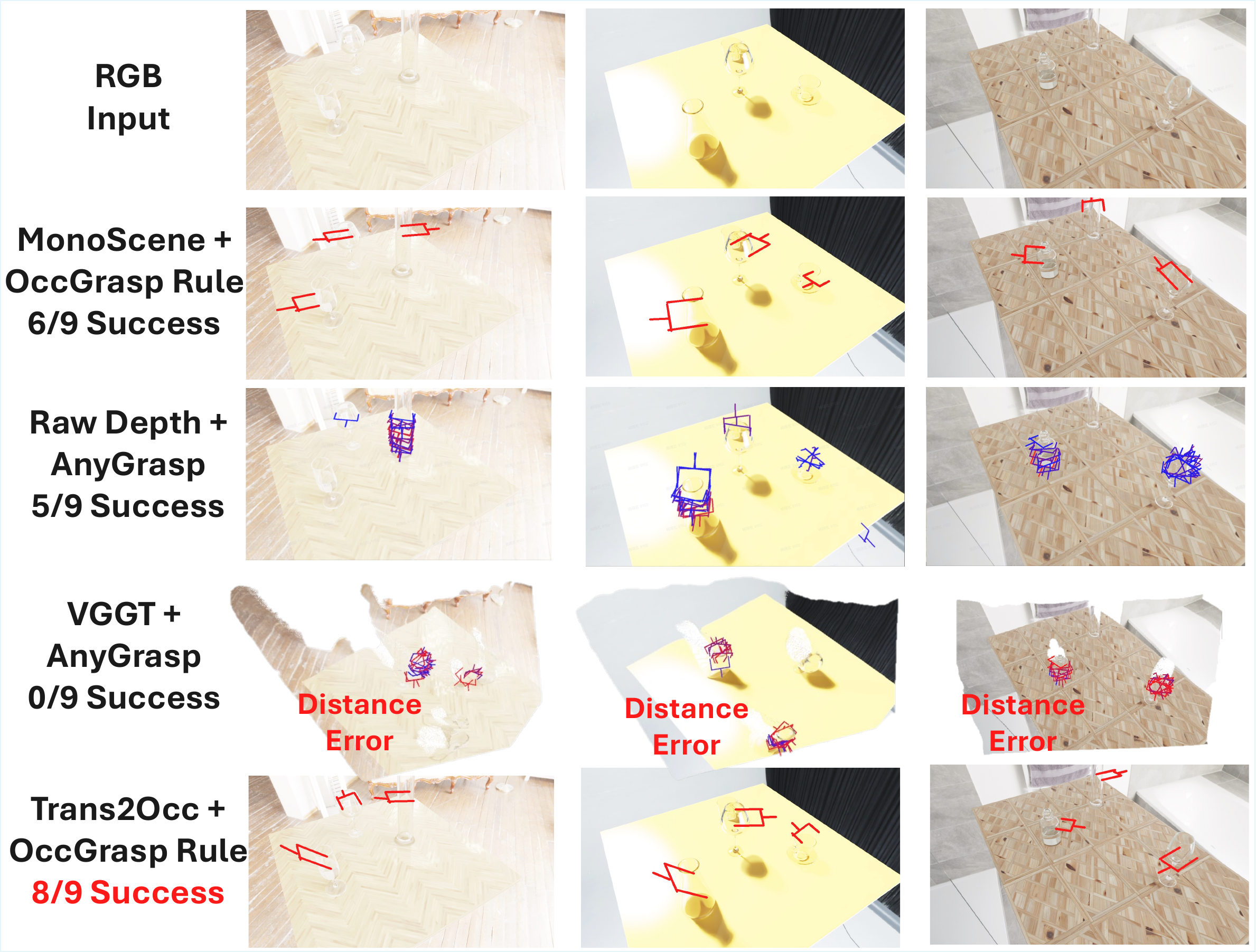}
    \caption{Comparison of grasp in simulation environment.Please zoom in to see the detail.}
    \label{fig:sim_grasp}
    \vspace{-2em}
\end{figure}

\begin{table}[t]
\centering
\small                         
\setlength{\tabcolsep}{3pt}    
\renewcommand{\arraystretch}{0.9} 
\caption{Grasp success rate comparison in the simulation.}
\resizebox{\linewidth}{!}{%
\begin{tabular}{lcc}
\toprule
Method & Success Rate (\%) & \# Trials \\
\midrule
VGGT + AnyGrasp    & 0.00  & 0 / 300 \\
GT Depth + AnyGrasp      & 44.10 & 132 / 300 \\
MonoScene + Rule-based   & 58.33 & 175 / 300 \\
\midrule
Trans2Occ + Rule-based        &\cellcolor[rgb]{ .706,  .776,  .906} 71.67 &\cellcolor[rgb]{ .706,  .776,  .906} 215 / 300 \\
\bottomrule
\end{tabular}}
\label{tab:grasp_success}
\vspace{-1em}
\end{table}

\subsubsection{In Real-world Environment}

\noindent \textbf{Occupancy Prediction and Visualization}
We directly deploy the model trained on Sim-Trans3D to real transparent-object scenes without fine-tuning. 
Given a single RGB image captured by a robot, Trans2Occ reconstructs consistent 3D occupancy and semantic volumes that align well with the real-world layouts. 
As shown in Figure~\ref{fig:occ_results}, the visualized results confirm that the learned occupancy representation generalizes robustly to real lighting and material variations.

\noindent \textbf{Grasp in Real-World}
To verify the performance of our method in simulation to real-world settings, we conduct real robot transparent object grasping experiments. 
For each experiment, a set of transparent objects is randomly placed on the table. 
The robot attempts grasps iteratively and stops when an object fails three times in a row. 
We evaluate performance using success rate, and  completion rate, defined as $\frac{\#\textrm{objects successfully grasped}}{\#\textrm{objects initially present}}$.
Table \ref{tab:real-experiment} reports the results, which shows the effectiveness and feasibility of our method. 
The captured images are fed directly into the trained Trans2Occ model without any additional processing.
The predicted occupancy is then used by the real robot to execute grasps on transparent objects such as beakers, flasks, and cylinders.
Despite being trained purely in simulation, the model achieves a high grasp success rate, demonstrating effective sim-to-real generalization and confirming that occupancy-based reconstruction is a reliable intermediate representation for transparent-object grasping.
We empirically found that simple affine warping and histogram equalization were sufficient for aligning real and synthetic images, without the need for explicit camera calibration.
\begin{figure}[t] 
    \centering
    \includegraphics[width=1\columnwidth]{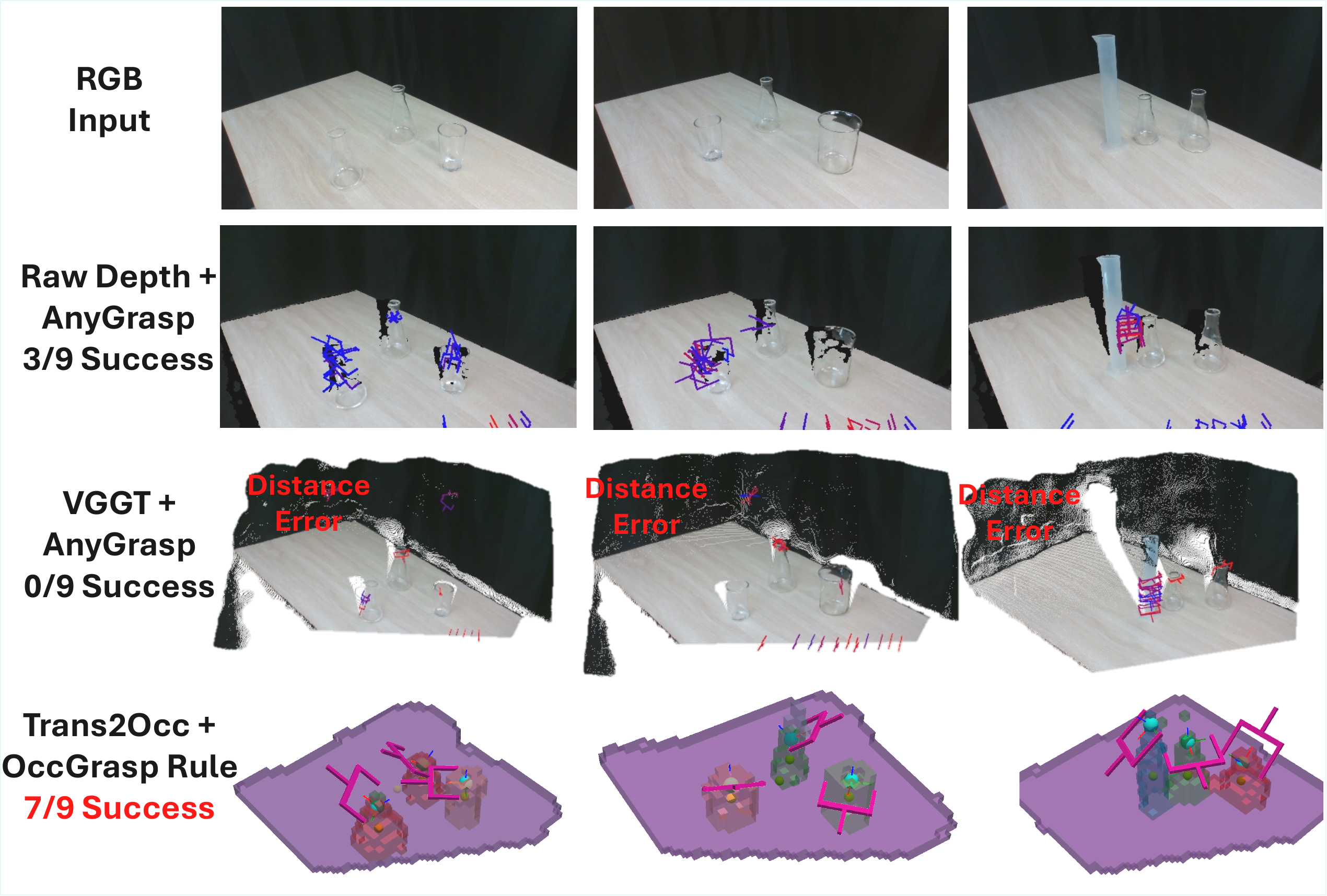}
    \caption{Comparison of grasp in real-world environment.}
    \label{fig:real_grasp}
\end{figure}

\begin{table}[t]
    \begin{center}
    \caption{Results of Real Robot Experiments}
    \label{tab:real-experiment}
    \resizebox{1\linewidth}{!}{
    \begin{tabular}{ccccc} \toprule
     & \textbf{\# Objects} & \textbf{\# Attempts} & \textbf{Success Rate} & \textbf{Completion Rate} \\ \midrule
    \textbf{Exper. 1} & 3& 9 & 66.7\% & 100.0\% \\ 
    \textbf{Exper. 2} & 3 & 9 & 55.6\% & 66.7\% \\ 
    \textbf{Exper. 3} & 3 & 9 & 44.4\% & 66.7\% \\ 
    \textbf{Exper. 4} & 3 & 8 & 62.5\% & 100.0\% \\ 
    \midrule
    \textbf{Ours Total} & 12 & 35 &\cellcolor[rgb]{ .706,  .776,  .906} 57.1\% &\cellcolor[rgb]{ .706,  .776,  .906} 83.3\% \\ 
    \textbf{RAW+AnyGrasp} & 12 & 35 & 28.6\% & 41.6\% \\
    \textbf{VGGT+AnyGrasp} & 12 & 35 & 5.7\% & 8.3\% \\ 

    \bottomrule
    \end{tabular}}
    \end{center}
    \vspace{-0.6cm}
\end{table}

\subsection{Ablation Studies}
In our ablation study, we evaluate the contribution of different components to the performance of our model. As shown in the previous experiments, grasping performance is strongly correlated with the quality of our occupancy reconstruction. Therefore, in the ablation analysis, we use the IoU of three representative categories: Beaker, Wine, and Cylinder, in the generalized setting as the primary metric.

\noindent \textbf{Data augmentation.}
Following our proposed augmentation strategy, we train models using three augmentation ratios and compare their effects on occupancy reconstruction. As shown in~\cref{tab:ablation_aug_branch}(a), moderate augmentation improves generalization. An augmentation ratio of 0.5 yields clear gains across Beaker, Cylinder, and Bottle. A very high augmentation ratio (1.0) leads to performance degradation, likely due to over-randomization.
Our final model uses a ratio of 0.7 and achieves the best overall performance, outperforming all fixed augmentation settings.

\noindent \textbf{Only 2D Branch or Geometry Branch.}
Our full method combines both the 2D branch and the Geometry branch to predict voxel occupancy. To assess the effect of each branch individually, we first test a variant that uses only the 2D branch. This setting reduces our method to a MonoScene-style approach and serves as a natural comparison point. We also evaluate a variant that uses only the Geometry branch. As shown in~\cref{tab:ablation_aug_branch}(b), each branch alone leads to a clear drop in performance, while the combination of both branches yields the best results.

\noindent \textbf{Parameters for grasping.}
We further analyze how each component of our grasping strategy contributes to overall stability. 
Table~\ref{tab:grasp_ablation} compares four settings: our full pipeline 
(32 candidate poses with PCA-based closing direction, collision checking, and angle scoring), 
removing the angle scoring term, removing non-target collision checking, 
and replacing the PCA-based closing direction with a random direction. 
The full version achieves the highest grasp success rate and the lowest penetration, 
while removing either geometric priors or collision constraints leads to clear degradation. 
This confirms that each module is necessary for robust and collision-free grasping.
\begin{table}[t]
\centering
\small                         
\setlength{\tabcolsep}{8pt}    
\renewcommand{\arraystretch}{0.9} 
\caption{Ablation study on data augmentation and branch design.}

\begin{tabular}{lccc}
\toprule
Configuration & Beaker & Cylinder & Bottle \\
\midrule
\multicolumn{4}{l}{\textit{(a) Effect of data augmentation ratio}} \\
Aug. ratio 0.0 & 42.24 & 38.67 & 48.31 \\
Aug. ratio 0.5 & 44.45 & 40.26 & 49.07 \\
Aug. ratio 1.0 & 43.19 & 39.42 & 47.52 \\
Ours (0.7)     & \cellcolor[rgb]{ .706,  .776,  .906}44.74 & \cellcolor[rgb]{ .706,  .776,  .906}39.57 & \cellcolor[rgb]{ .706,  .776,  .906}49.89 \\
\midrule
\multicolumn{4}{l}{\textit{(b) Effect of 2D and 3D branches}} \\
Only 2D branch   & 35.38 & 36.75 & 47.51 \\
Only Geometry branch   & 32.52 & 27.86 & 34.17 \\
Ours (2D+Geometry)     & \cellcolor[rgb]{ .706,  .776,  .906}44.74 & \cellcolor[rgb]{ .706,  .776,  .906}39.57 & \cellcolor[rgb]{ .706,  .776,  .906}49.89 \\
\bottomrule
\end{tabular}
\label{tab:ablation_aug_branch}
\end{table}



\begin{table}[t]
\centering
\small                         
\setlength{\tabcolsep}{8pt}    
\renewcommand{\arraystretch}{0.9} 
\caption{Ablation study on the grasping policy.}
\begin{tabular}{lcc}
\toprule
Setting & SR (\%) & Penetration (cm) \\
\midrule
w/o angle scoring & 20.72 & 0.42 \\
w/o  collision checking & 51.53 & 0.93 \\
Random closing direction & 43.82 & 0.47 \\
\midrule
Ours & \cellcolor[rgb]{ .706,  .776,  .906}71.67 &\cellcolor[rgb]{ .706,  .776,  .906} 0.14 \\
\bottomrule
\end{tabular}
\label{tab:grasp_ablation}
    \vspace{-0.6cm}
\end{table}

\section{Conclusion}

We presented Trans2Occ, a practical single-view framework for transparent object perception and manipulation based on voxel occupancy prediction. To enable scalable supervision, we introduced Sim-Trans3D, a simulation pipeline that generates paired RGB observations and voxel-level annotations for diverse transparent objects.
By learning a direct mapping from RGB images to voxel occupancy, our approach produces geometry-consistent 3D representations that effectively support downstream grasp planning. Experimental results demonstrate accurate occupancy prediction and reliable grasp execution in both simulation and real-world environments. Notably, models trained purely on synthetic data generalize to real-world setups without additional adaptation, highlighting the robustness of voxel occupancy as a domain-invariant intermediate representation.
Overall, our findings suggest that single-view occupancy prediction provides a practical and scalable solution for transparent object manipulation. Future work includes incorporating physical interaction to resolve geometric ambiguities and extending the framework to more complex scenarios such as transparent containers with dynamic contents.

{
\small
\bibliographystyle{IEEEtran}
\bibliography{main}
}





\end{document}